\documentclass{INTERSPEECH2023}
\usepackage{bm}
\usepackage{cite}

\interspeechcameraready

\title{End-to-End Joint Target and Non-Target Speakers ASR}
\name{Ryo Masumura, Naoki Makishima, Taiga Yamane, Yoshihiko Yamazaki, Saki Mizuno, Mana Ihori, Mihiro Uchida, Keita Suzuki, Hiroshi Sato, Tomohiro Tanaka, Akihiko Takashima, Satoshi Suzuki, Takafumi Moriya, Nobukatsu Hojo, Atsushi Ando}
\address{NTT Computer and Data Science Laboratories, NTT Corporation, Japan}
\email{ryo.masumura@ntt.com}

\begin{document}

\maketitle
 
\begin{abstract}
This paper proposes a novel automatic speech recognition (ASR) system that can transcribe individual speaker's speech while identifying whether they are target or non-target speakers from multi-talker overlapped speech. Target-speaker ASR systems are a promising way to only transcribe a target speaker's speech by enrolling the target speaker’s information. However, in conversational ASR applications, transcribing both the target speaker’s speech and non-target speakers’ ones is often required to understand interactive information. To naturally consider both target and non-target speakers in a single ASR model, our idea is to extend autoregressive modeling-based multi-talker ASR systems to utilize the enrollment speech of the target speaker. Our proposed ASR is performed by recursively generating both textual tokens and tokens that represent target or non-target speakers. Our experiments demonstrate the effectiveness of our proposed method.
\end{abstract}

\section{Introduction}
Single-talker automatic speech recognition (ASR) systems limit their applicability because multiple utterances are often overlapped in many practical scenes. Therefore, recognizing multi-talker overlapped monaural speech signals has been the focus of much attention \cite{park_csl2022,zmolikova_arxiv2023}. In particular, target-speaker ASR (TS-ASR) systems that can transcribe only a target speaker's speech from the overlapped ones is known to be an effective way when the target speaker is specified. The systems are informed about the target speaker using an enrollment speech of that speaker, and can extract the same speaker's speech as the enrollment one. In this paper, we aim to improve ASR systems that use the speaker enrollment to make it suitable for conversational ASR applications such as meetings.

Toward a better performance of TS-ASR, several methods have been developed with the progress of deep learning technology. Initial TS-ASR systems have been built by cascading single-talker ASR with target-speaker speech extraction \cite{delcroix_2018,wang_interspeech2019,xu_asru2019,wang_interspeech2020,ji_icassp2020}, which is a process to separate a target speaker's speech from a multi-talker overlapped one. The cascading systems are comparatively feasible to build because individual components are separately trained, but their performance is limited because overall optimization of TS-ASR cannot be achieved. Therefore, previous studies challenge to directly build TS-ASR in an end-to-end manner \cite{delcroix_interspeech2019,denisov_interspeech2019,moriya_interspeech2022}. The main advantage of the end-to-end TS-ASR systems is that overall optimization to transcribe a target speaker's speech into text can be performed. In addition, since there is no need to separate the target speaker's speech as a signal, computation complexity can be reduced compared with the cascading systems.

TS-ASR systems are really effective to execute personalized speech commands in a cocktail-party situation or record personalized life-logs; however, ignoring non-target speakers' speech is unsuitable for understanding interactions between the target speaker and non-target speakers (e.g., interactions between a target salesperson and customers in a store). In conversational ASR applications, both the target speaker's speech and non-target speakers’ ones often need to be transcribed. In addition, we suspect that eliminating non-target speakers' speech is ineffective to perform accurate TS-ASR. Considering non-target speakers' information will explicitly enables us to model the dependency among the speakers and lead to improving TS-ASR performance. In fact, considering non-target interference speaker's information has achieved better performance in senone-based acoustic modeling \cite{kanda_interspeech2019}. One challenge is how we naturally consider both the target speaker and non-target speakers in an end-to-end ASR framework.

In this paper, we propose end-to-end joint target and non-target speakers ASR (TS-NTS-ASR) systems that jointly transcribe a target speaker's and non-target speakers’ speech using a unified autoregressive model. Our key idea is to extend autoregressive modeling-based multi-talker ASR (MT-ASR) systems (detailed in Sections 2 and 3.2) \cite{kanda_interspeech2020a,masumura_interspeech2021} to utilize the enrollment speech of the target speaker. In existing MT-ASR systems, transcriptions of multiple speakers are generated as a serialized token sequence, but it does not determine whether each transcription is spoken by the target speaker or not. Therefore, our proposed method utilizes one unified autoregressive model for not only transcribing all speakers' speech but also identifying whether they are the target speaker or non-target speakers using the enrollment speech. The identification is achieved by handling both textual tokens and tokens that represent the target or non-target speakers within the serialized token sequence. Our proposed TS-NTS-ASR systems have two advantages: they can transcribe both the target speaker's speech and non-target speakers' speech in a simple beam search decoding, and they improve TS-ASR performance compared with conventional TS-ASR systems that ignore non-target speakers' speech.

We detail our contributions as follows.
\begin{itemize}
\item This paper is the first study of end-to-end joint target and non-target speakers ASR systems. For the proposed autoregressive modeling, we show three serialized token sequence patterns to take the target speaker and non-target speakers' spoken text into consideration. We also show a network structure that can effectively utilize the speaker enrollment in autoregressive modeling.
\item This paper also shows a method to implement non-target speakers ASR (NTS-ASR) systems that transcribe an individual speaker's speech except for the target speaker's speech.
\item This paper demonstrates that our proposed method effectively transcribes both the target speaker's speech and non-target speakers' speech in experiments using a Japanese multi-talker overlapped speech dataset.
\end{itemize}

\section{Related Work}

\smallskip
\noindent {\bf Speech processing with target speaker enrollment:}
This paper is related to speech processing that uses speaker enrollment. In previous studies, target-speaker speech extraction \cite{delcroix_2018,wang_interspeech2019,xu_asru2019,wang_interspeech2020,ji_icassp2020}, TS-ASR \cite{delcroix_interspeech2019,denisov_interspeech2019,moriya_interspeech2022}, and target-speaker voice activity detection \cite{medenniko_interspeech2020,ding_odyssey2020,he_interspeech2021,makishima_interspeech2021,ding_interspeech2022,wang_icassp2022} have been mainly investigated. These studies use the enrollment speech of a target speaker to extract the target speaker's information. In contrast, our proposed method examines speech processing for both the target speaker and non-target speakers using the target speaker enrollment.

\smallskip
\noindent {\bf Multi-talker ASR:}
This paper is related to end-to-end MT-ASR systems that can transcribe an individual speakers' speech from a multi-talker overlapped one. Initial end-to-end MT-ASR systems introduce modeling with multiple output branches, in which each branch generates a transcription for one speaker by considering all possible permutations of speakers \cite{yu_interspeech2017,chang_icassp2019,seki_acl2018,settle_icassp2018,chang_icassp2020,sklyar_arxiv2020,tripathi_2020}. A recent hopeful approach for end-to-end modeling is autoregressive modeling in which transcriptions of multiple speakers are recursively generated from one output branch \cite{kanda_interspeech2020a,masumura_interspeech2021}. Our proposed method is regarded as an extended modeling of the latter MT-ASR to identify a target speaker or non-target speakers using the target speaker enrollment.

\section{Preliminaries}
This section describes target-speaker ASR (TS-ASR) and multi-talker ASR (MT-ASR) systems based on autoregressive modeling. These two systems can handle monaural multi-talker overlapped speech. The former generates only the target speaker's spoken text and the latter generates all speakers' spoken text.

\subsection{Target-speaker ASR}
TS-ASR based on autoregressive modeling predicts a generation probability of a target speaker's spoken text $\bm{W}=\{w_1,\cdots,w_N\}$ from monaural multi-talker overlapped speech $\bm{X}$ and the target speaker's enrollment speech $\bm{E}$, where $w_n \in {\cal V}$ is the $n$-th token in the spoken text, $N$ is the number of tokens in the spoken text, and ${\cal V}$ is the vocabulary set. In autoregressive modeling, the generation probability of $\bm{W}$ is defined as
\begin{equation}
  P(\bm{W}|\bm{X}, \bm{E}; \bm{\Theta}_{\tt ts}) = \prod_{n=1}^N P(w_n|w_{1:n-1}, \bm{X}, \bm{E}; \bm{\Theta}_{\tt ts}) ,
\end{equation}
where $\bm{\Theta}_{\tt ts}$ represents the trainable model parameter sets and $w_{1:n-1}=\{w_1,\cdots,w_{n-1}\}$. The loss function to optimize the model parameter sets is defined as
\begin{equation}
  {\cal L}(\bm{\Theta}_{\tt ts}) = - \sum_{(\bm{X},\bm{E},\bm{W}) \in {\cal D}_{\tt ts}} \log P(\bm{W}|\bm{X},\bm{E}; \bm{\Theta}_{\tt ts}) , 
\end{equation}
where ${\cal D}_{\tt ts}$ represents a paired dataset of input multi-talker speech, the target speaker's enrollment speech, and the target speaker's spoken text. Note that spoken text becomes empty when the target speaker is not included in the multi-talker overlapped speech.

\subsection{Multi-talker ASR}
MT-ASR based on autoregressive modeling predicts a generation probability of all speakers' spoken text $\bm{W}^{1:T} = \{\bm{W}^1,\cdots,\bm{W}^T\}$ from monaural multi-talker overlapped speech $\bm{X}$, where $\bm{W}^t=\{w_1^t,\cdots,w_{N^t}^t\}$ is the $t$-th speaker's spoken text, $N^t$ is the number of tokens in the $t$-th speaker's spoken text, and $T$ is the number of speakers in the multi-talker overlapped speech. Multiple spoken texts in one autoregressive model are serialized into a single token sequence. Thus, the generation probability of $\bm{W}^{1:T}$ is defined as
\begin{equation}
  \begin{split}
  P(\bm{W}^{1:T} |\bm{X}; \bm{\Theta}_{\tt mt}) & =  P(\bm{S} |\bm{X}; \bm{\Theta}_{\tt mt})  \\
  & = \prod_{l=1}^{|\bm{S}|} P(s_l|s_{1:l-1}, \bm{X}; \bm{\Theta}_{\tt mt}) ,
  \end{split}
\end{equation}
where $\bm{\Theta}_{\tt mt}$ represents the trainable model parameter sets, $\bm{S} = \{s_1,\cdots,s_{|\bm{S}|}\}$ is the serialized token sequence, and $s_l \in \{{\cal V} \cup {\cal O}\}$ is the $l$-th token in the serialized token sequence. $\cal O = \{{\tt [sep]}, {\tt [eos]}\}$ represents the special token set, where $\tt [sep]$ represents the speaker change and $\tt [eos]$ represents the end-of-sentence. There are multiple permutations in the order of the multiple spoken texts $\bm{W}^{1:T}$, so they are sorted by their start times, which is called first-in first-out. When the speaker index $t$ is ordered by the start time, the serialized token sequence is represented as
\begin{multline}
  \bm{S} = \{w_1^1,\cdots,w_{N^1}^1, {\tt [sep]}, w_1^2,\cdots,w_{N^2}^2,  \\
  \cdots, w_{N^{T-1}}^{T-1},{\tt [sep]}, w_1^T,\cdots,w_{N^T}^T, {\tt [eos]}\} .
\end{multline}
Thus, the serialized token sequence is composed by concatenating multiple spoken texts while inserting $\tt [sep]$ between them and $\tt [eos]$ at the end of the entire sequence.

The loss function to optimize the model parameter sets is defined as
\begin{equation}
  {\cal L}(\bm{\Theta}_{\tt mt}) = - \sum_{(\bm{X},\bm{S}) \in {\cal D}_{\tt mt}}  \log  P(\bm{S}|\bm{X}; \bm{\Theta}_{\tt mt}) , 
\end{equation}
where ${\cal D}_{\tt mt}$ represents a paired dataset of the serialized token sequence and the multi-talker overlapped speech.

\section{Proposed Method}
This paper proposes end-to-end joint target and non-target speaker ASR (TS-NTS-ASR) systems that can transcribe an individual speaker's speech while identifying whether they are the target speaker or non-target speakers from multi-talker overlapped speech and the target speaker's enrollment speech.

\subsection{Modeling}
We construct a TS-NTS-ASR model by extending autoregressive modeling-based MT-ASR (described in Section 3.2) to utilize the target speaker's enrollment speech. The TS-NTS-ASR model predicts a joint generation probability of a target speaker's spoken text $\bm{W}=\{w_1,\cdots,w_N\}$ and non-target speakers' spoken texts $\bar{\bm{W}}^{1:T-1} = \{\bar{\bm{W}}^1,\cdots,\bar{\bm{W}}^{T-1}\}$ from monaural multi-talker overlapped speech $\bm{X}$ and the target speaker's enrollment speech $\bm{E}$, where $\bar{\bm{W}}^t=\{\bar{w}_1^t,\cdots,\bar{w}_{N^t}^t\}$ is the $t$-th non-target speaker's spoken text. To handle these outputs within an autoregressive model, the target speaker's spoken text and non-target speakers' spoken texts are serialized as a single token sequence $\bm{Z}=\{z_1,\cdots,z_{|\bm{Z}|}\}$. In this case, the joint generation probability is defined as
\begin{equation}
  \begin{split}
  P(\bm{W},\bar{\bm{W}}^{1:T-1} |\bm{X}, \bm{E}; \bm{\Theta}) & =  P(\bm{Z} |\bm{X}, \bm{E}; \bm{\Theta})  \\
  & = \prod_{l=1}^{|\bm{Z}|} P(z_l|z_{1:l-1}, \bm{X}, \bm{E}; \bm{\Theta}) ,
  \end{split}
\end{equation}
where $\bm{\Theta}$ represents the trainable model parameter sets. $z_l \in \{{\cal V} \cup {\cal U}\}$ is the $l$-th token in the serialized token sequence. $\cal U = \{{\tt [t]}, {\tt [nt]}, {\tt [eos]}\}$ represents the special token set, where $\tt [t]$ represents the target speaker's section and $\tt [nt]$ represents non-target speaker's section.

TS-NTS-ASR systems can be associated with those described in Section 3. By disregarding $\bm{E}$ from Eq. (6) to not identify the target or non-target speakers, the system can be regarded as the same modeling as MT-ASR defined by Eq. (3). In addition, by disregarding $\bar{\bm{W}}^{1:T-1}$ from Eq. (6), the system can be regarded as the same modeling as TS-ASR defined by Eq. (1). On the other hand, by disregarding $\bm{W}$ from Eq. (6), the system can be regarded as a non-target speakers ASR (NTS-ASR) system that transcribes an individual speaker's speech except for the target speaker's speech.

The loss function to optimize the model parameter sets is defined as  
\begin{equation}
  {\cal L}(\bm{\Theta}) = - \sum_{(\bm{X},\bm{E},\bm{Z}) \in {\cal D}}  \log  P(\bm{Z}|\bm{X},\bm{E}; \bm{\Theta}) , 
\end{equation}
where ${\cal D}$ represents a paired dataset of input multi-talker speech, the target speaker's enrollment speech, and the serialized token sequence. 

\subsection{Serialization of tokens}
For the proposed autoregressive modeling, we show three serialized token sequence patterns. In each serialization, we can transcribe an individual speaker’s speech while identifying whether they are the target or non-target speaker.

\smallskip
\noindent {\bf Target-speaker first:}
We put the target speaker's spoken text first, then the non-target speakers' spoken texts. The serialized token sequence is composed by
\begin{multline}
  \bm{Z} = \{{\tt [t]},w_1,\cdots,w_{N},
  {\tt [nt]}, \bar{w}_1^1,\cdots,\bar{w}_{N^1}^1,   \\
  \cdots, {\tt [nt]},\bar{w}_1^{T-1},\cdots,\bar{w}_{N^{T-1}}^{T-1}, {\tt [eos]}\} .
\end{multline}
Note that we can perform TS-ASR by stopping when the first ${\tt [nt]}$ is generated.

\smallskip
\noindent {\bf Non-target speaker first:}
We put the non-target speakers' spoken texts first, then the target speaker's spoken text. The serialized token sequence is composed by
\begin{multline}
  \bm{Z} = \{{\tt [nt]}, \bar{w}_1^1,\cdots,\bar{w}_{N^1}^1,\cdots,{\tt [nt]}, \bar{w}_1^{T-1},\cdots,\bar{w}_{N^{T-1}}^{T-1}, \\
  {\tt [t]},w_1,\cdots,w_{N}, {\tt [eos]}\} .
\end{multline}
Note that we can perform NTS-ASR by stopping when the first ${\tt [t]}$ is generated.

\smallskip
\noindent {\bf First-in first-out:}
We sort each transcription by their start times regardless of whether the text is spoken by the target or non-target speaker. For example, when the target speaker starts speaking a second earlier among all speakers, the serialized token sequence is composed by
\begin{multline}
  \bm{S} = \{{\tt [nt]},\bar{w}_1^1,\cdots,\bar{w}_{N^1}^1,{\tt [t]}, w_1,\cdots,w_{N}, \\
  {\tt [nt]},\bar{w}_1^2,\cdots,\bar{w}_{N^2}^2, \cdots,{\tt [nt]}, \bar{w}_1^{T-1},\cdots,\bar{w}_{N^{T-1}}^{T-1}\}.
\end{multline}
Note that this is the same arrangement as that for MT-ASR.

\begin{figure}[t]
  \begin{center}
    \includegraphics[width=80mm]{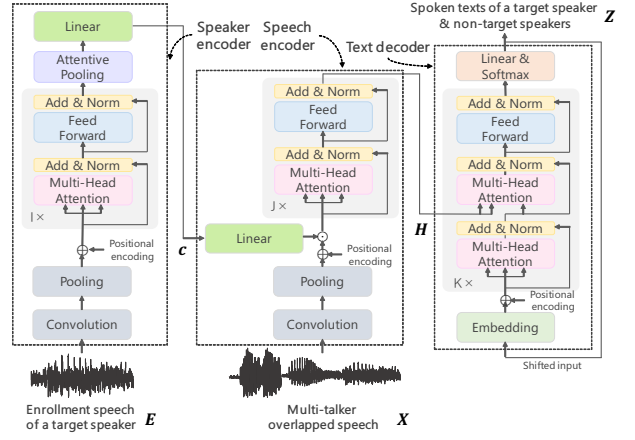}
  \end{center}
  \vspace{-5mm}
  \caption{Network structure of TS-NTS-ASR.}
  \vspace{-4mm}
\end{figure}

\begin{table*}[t!]
  \begin{center}
      \caption{Experimental results in terms of character error rate (\%).}
      \vspace{-3mm} 
      \begin{tabular}{|lcc|rr|rrr|} \hline
      System &Speaker& Serialization &\multicolumn{2}{c|}{Single-talker speech} & \multicolumn{3}{c|}{Multi-talker overlapped speech} \\
      &enrollment & of tokens & Target & Non-target & Target & Non-target & Multi-talker  \\ \hline \hline
        MT-ASR & - & First-in first-out  & - & - &  - & - & 12.75 \\ \hline \hline
        TS-ASR & $\checkmark$ & - & 6.95 & -& 12.54 & - & - \\ 
        NTS-ASR & $\checkmark$ & - & - & 8.36 & - & 13.80 & - \\    
        TS-NTS-ASR (proposed) & $\checkmark$ & Target first & 6.15 & 7.87 & 12.36 & 13.05 & 12.60 \\ 
        TS-NTS-ASR (proposed) & $\checkmark$ & Non-target first & 6.52 & 7.40 & 13.18 & 13.24 & 13.57 \\ 
       TS-NTS-ASR (proposed) & $\checkmark$ & First-in first-out & {\bf 5.29} & {\bf 7.14} & {\bf 11.80} & {\bf 12.53} & {\bf 12.15} \\ \hline
      \end{tabular}
  \end{center}
  \vspace{-6mm}
\end{table*}

\subsection{Network structure}
We construct the TS-NTS-ASR system from two speaker encoders and a text decoder. Figure 1 shows the network structure of TS-NTS-ASR.

\smallskip
\noindent {\bf Speaker encoder:} The speaker encoder converts acoustic features of the target speaker's enrollment speech into a speaker vector. This conversion is defined as
\begin{equation}
\bm{c} = {\rm SpeakerEnc}(\bm{E}; \bm{\theta}_{\rm e}), 
\end{equation}
where $\bm{\theta}_{\rm e} \in \bm{\Theta}$ is its parameters. We implement this function using convolution and pooling layers, a positional encoding layer, transformer encoder blocks, an attentive pooling layer, and a linear layer.

\noindent {\bf Speech encoder:} The speech encoder converts acoustic features of the multi-talker overlapped speech and the speaker vector into hidden representations. This conversion is defined as
\begin{equation}
\bm{H} = {\rm SpeechEnc}(\bm{X},\bm{c}; \bm{\theta}_{\rm x}), 
\end{equation}
where $\bm{\theta}_{\rm e} \in \bm{\Theta}$ is its parameters. We implement this function using convolution and pooling layers to generate subsampled speech representations, a positional encoding layer, an element-wise multiplication layer with a linear layer, and transformer encoder blocks. The element-wise multiplication is performed between the speaker vector and subsampled speech representations to consider the target speaker's information \cite{xu_asru2019}.

\noindent {\bf Text decoder:} The token decoder computes the generation probability of a token given preceding tokens and the hidden representations produced in the speech encoder. The generation probability is computed from
\begin{equation}
P(z_l|z_{1:l-1},\bm{X},\bm{E},\bm{\Theta}) = {\rm TextDec}(z_{1:l-1},\bm{H}; \bm{\theta}_{\rm z}),
\end{equation}
where $\bm{\theta}_{\rm z} \in \bm{\Theta}$ is its parameters. We implement this function using a linear embedding layer, a positional encoding layer, transformer decoder blocks, and a softmax layer with a linear transformation.

\section{Experiments}
In the experiments, we used the Corpus of Spontaneous Japanese (CSJ) \cite{maekawa_lrec2000}, which is a single-talker speech dataset. We first split the dataset into training set (518.4 h), validation set (1.3 h), and test set (1.9 h) and, following procedures were conducted for each set. To generate multi-talker overlapped speech dataset, we mixed multiple audio signals as a monaural signal. To this end, we randomly chose multiple audio signals so as not to select the same speakers. We set the number of speakers in the mixed signals as two or three. When mixing the audio signals, the original volume of each utterance was kept unchanged, resulting in an average signal-to-interference ratio of about 0 dB. For the delay applied to each utterance, the start times of the individual utterances differ by 0.5 s or longer. In addition, every utterance in each mixed audio sample has at least one speaker-overlapped region with other utterances. In addition, we prepared enrollment speech for the single-talker speech and the multi-talker overlapped speech sets. For the single-talker speech dataset, we prepared two cases: a different utterance of the target speaker was used as the speaker enrollment, and an utterance of a different speaker from the target speaker was used as the speaker enrollment. For the multi-talker overlapped speech dataset, an utterance spoken by one of the talkers was used as the enrollment. In training, all paired patterns (about 2,500 h) were jointly used. In testing, each of paired patterns were individually evaluated.

\subsection{Setups}
We constructed an MT-ASR, TS-ASR, NTS-ASR, and three TS-NTS-ASR systems. Except for MT-ASR, which does not use speaker enrollment, we introduced the network structure described in Section 4.3 for each system. MT-ASR used a network structure that excludes a speaker encoder and its connection. For these systems, the transformer blocks were composed under the following conditions: the dimensions of the output continuous representations were set to 512, the dimensions of the inner outputs were set to 2,048, and the number of heads in the multi-head attentions was set to 4. In the nonlinear transformational functions, the Swish activation was used. For both the speaker and the speech encoders, we used 80 log mel-scale filterbank coefficients as acoustic features. The frame shift was 10 ms. The acoustic features passed two convolution and max pooling layers with a stride of 2, so we down-sampled them to $1/4$ along with the time axis. After these layers, we stacked 4-layer transformer encoder blocks. For the text decoder, we used 512-dimensional character embeddings where the vocabulary size was set to 3,262. We also stacked 3-layer transformer decoder blocks. For the training, we used the RAdam optimizer \cite{liu_iclr2020}. For the systems that used speaker enrollment, the speaker encoder was first trained using VoxCeleb2 dataset \cite{chung_interspeech2018} with ArcFace criterion \cite{deng_cvpr2018}, and then the speech encoder and text decoder were trained while freezing the speaker encoder. We set the mini-batch size to 64 utterances and the dropout rate in the transformer blocks to 0.1. We introduced label smoothing where its smoothing parameter was set to 0.1. In addition, we applied SpecAugment \cite{park_is2019} and enrollment-less training \cite{makishima_interspeech2021}. For testing, we used a beam search algorithm in which the beam size was set to 4.

\subsection{Results}
Table 1 shows the results in terms of character error rate for each ASR system. The ``Target'' column represents TS-ASR performance and the ``Non-target'' column represents NTS-ASR performance, each of which was only evaluated when using the speaker enrollment. The ``Multi-talker'' column represents MT-ASR performance that evaluated all speakers' results. First, the results show that each TS-NTS-ASR system outperformed the TS-ASR system. This indicates that considering non-target speakers' information explicitly leads to improving TS-ASR performance. In addition, each TS-NTS-ASR system outperformed the NTS-ASR system. This also indicates that joint consideration of the target and non-target speakers is important. Next, among the TS-NTS-ASR systems, the best results were attained by ``first-in first-out'' serialization. It is considered that ``first-in first-out,'' which can attend to the input of the multi-talker overlapped speech in left-to-right order, is easier to train than ``target first'' and ``non-target first.'' Furthermore, Table 2 shows the results of evaluating the target speaker or non-target speaker detection error rate against single-talker speech to reveal the performance of considering the speaker enrollment. The error was calculated using the number of utterances that misdetects the target or non-target speaker. The results show that TS-NTS-ASR could reduce the detection error rate for both the target and non-target speakers. This induces TS-ASR and NTS-ASR performance improvements. These results demonstrate that our proposed method effectively transcribes not only the target speaker's speech but also non-target speakers' speech.

\begin{table}[t!]
 \label{}
 \begin{center}
   \caption{Experimental results in terms of target speaker or non-target speaker detection error rate (\%). TS-NTS-ASR system introduced ``first-in first-out'' serialization.}
   \vspace{-3mm}
   \begin{tabular}{|l|rr|} \hline
    & \multicolumn{2}{c|}{Single-talker speech} \\
    & Target & Non-target \\ \hline \hline
    TS-ASR & 1.41 & 2.03 \\
    NTS-ASR & 2.10 & 2.35 \\
    TS-NTS-ASR (proposed) & {\bf 0.00} & {\bf 1.82} \\  \hline
  \end{tabular}
 \end{center}
 \vspace{-6mm}
\end{table}

\section{Conclusions}
We presented end-to-end joint target and non-target speakers ASR systems that jointly transcribe a target speaker's speech and non-target speakers’ speech using the target speaker's enrollment speech. The key strength of the proposed method is to effectively transcribe all speakers' speech while identifying whether they are the target speaker or non-target speakers using the enrollment speech. This is suitable to understand interactions between the target speaker and non-target speakers. Our experimental results showed that the proposed method effectively transcribes not only the target speaker's speech but also non-target speakers' speech, and outperformed the target-speaker ASR system and the non-target speaker ASR system.

\end{document}